\documentclass[letterpaper]{article} 
\usepackage{aaai24}  
\usepackage{times}  
\usepackage{helvet}  
\usepackage{courier}  
\usepackage[hyphens]{url}  
\usepackage{graphicx} 
\usepackage{natbib}  
\usepackage{caption} 
\usepackage{algorithm}
\usepackage{algorithmic}
\usepackage{bm}
\usepackage{amsmath}
\usepackage{amssymb}
\usepackage{cases}
\usepackage{multicol}
\usepackage{multirow}
\usepackage{graphicx}
\usepackage{booktabs}
\usepackage{color}
\usepackage{array}

\usepackage{newfloat}
\usepackage{listings}
\DeclareCaptionStyle{ruled}{labelfont=normalfont,labelsep=colon,strut=off} 
\floatstyle{ruled}
\newfloat{listing}{tb}{lst}{}
\floatname{listing}{Listing}

\author{
	Yide Qiu, Shaoxiang Ling, Tong Zhang,  Bo Huang, Zhen Cui$^{*}$
}
\affiliations{
	Nanjing University of Science and Technology, Nanjing, China\\
\{q115025886, lingshaoxiang, tong.zhang, huangbo, zhen.cui\}@njust.edu.cn}

\title{UniKG: A Benchmark and Universal Embedding for Large-Scale Knowledge Graphs}

\usepackage{bibentry}

\begin{document}

\maketitle

\begin{abstract}
Irregular data in real-world are usually organized as heterogeneous graphs (HGs) consisting of multiple types of nodes and edges. To explore useful knowledge from real-world data, both the large-scale encyclopedic HG datasets and corresponding effective learning methods are crucial, but haven't been well investigated. In this paper, we construct a large-scale HG benchmark dataset named UniKG from Wikidata to facilitate knowledge mining and heterogeneous graph representation learning. Overall, UniKG contains more than 77 million multi-attribute entities and 2000 diverse association types, which significantly surpasses the scale of existing HG datasets. To perform effective learning on the large-scale UniKG, two key measures are taken, including (i) the semantic alignment strategy for multi-attribute entities, which projects the feature description of multi-attribute nodes into a common embedding space to facilitate node aggregation in a large receptive field; (ii) proposing a novel plug-and-play anisotropy propagation module (APM) to learn effective multi-hop anisotropy propagation kernels, which extends methods of large-scale homogeneous graphs to heterogeneous graphs. These two strategies enable efficient information propagation among a tremendous number of multi-attribute entities and meantimes adaptively mine multi-attribute association through the multi-hop aggregation in large-scale HGs. We set up a node classification task on our UniKG dataset, and evaluate multiple baseline methods which are constructed by embedding our APM into large-scale homogenous graph learning methods. Our UniKG dataset and the baseline codes have been released at \url{https://github.com/Yide-Qiu/UniKG}. 
\end{abstract}

\section{Introduction}

Heterogeneous Graphs (HGs), also known as Heterogeneous Information Networks (HINs), consist of multiple types of nodes and links. Compared with homogeneous graphs, the heterogeneity in multi-attribute nodes and graph topology makes HGs carry richer semantics, and be more suitable to characterize a variety of complex real-world systems, such as academic networks and social networks. For this reason, methods that focus on representation learning on HGs have drawn increasing attention in recent years, and have facilitated numerous application tasks in diverse domains, including recommendation systems~\cite{gao2022graph, wu2022graph}, malware detection systems~\cite{zhang2022practical}, and healthcare systems~\cite{cao2022generalizability}.

\begin{figure}[t] 
	\centering
	\includegraphics[width=1\linewidth]{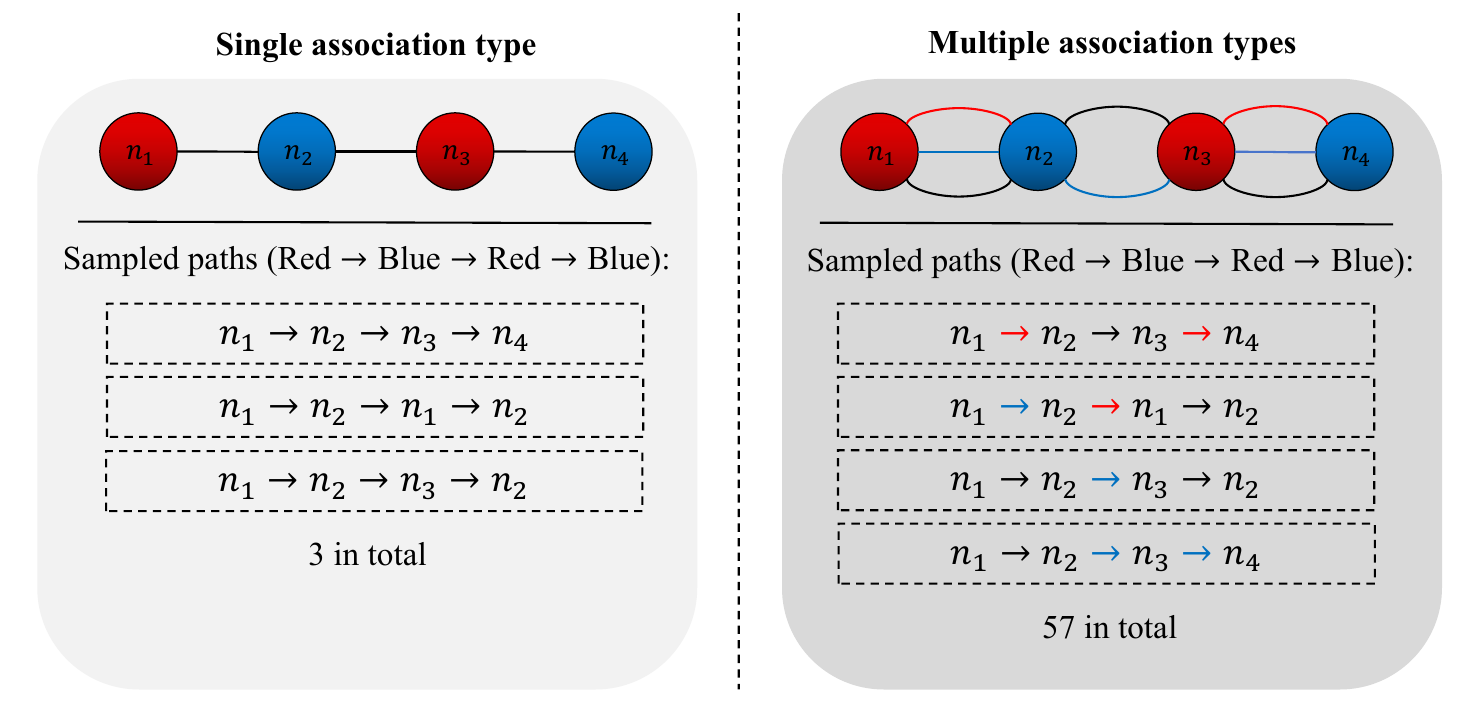} 
	\caption{An example of the meta-path method. There are two kinds of nodes (red and blue) in the graph, and the paths are sampled in the order of `red, blue, red, blue', starting from the node $n_1$. The different colors of the edges represent different types.}
	\label{fig:pathmethod}
\end{figure}

Existing HG-related works involve the design of effective learning methods and also the construction of HG datasets, because an encyclopedic HG dataset is also crucial to promote HG learning. Regarding HG dataset construction, existing research~\cite{sharma2022finred, chen2023knowledge, kubevsa2023damuel, whitehouse2023webie, ahrabian2023pubgraph, stranisci2022urw, luo2023dhge} mostly focuses on combining specific entities and relationships within knowledge graphs (KGs) to facilitate downstream tasks within the corresponding domain. Existing datasets generally consist of thousands of nodes and dozens of relationships applied in a specific domain such as finance, recommendation systems and social networks. For the learning methods on HGs, depending on the difference of methods in capturing semantic structure, the existing works can be broadly classified into aggregation-based methods~\cite{schlichtkrull2018modeling,zhang2019heterogeneous,xu2019relation} and meta-path-based methods~\cite{wang2020dynamic,Huang_Zheng_Cheng_Sun_Mamoulis_Li_2016,Sun_Han_Yan_Yu_Wu_2011,Chang_Chen_Hu_Zheng_Zhou_Chen_2022}. The aggregation-based methods primarily concentrate on iteratively aggregating neighbor information to update node representations. In contrast, meta-path-based methods focus on combining information from distinct meta-paths to learn node representations. 

\begin{figure*}[t] 
	\centering
	\includegraphics[width=1\linewidth]{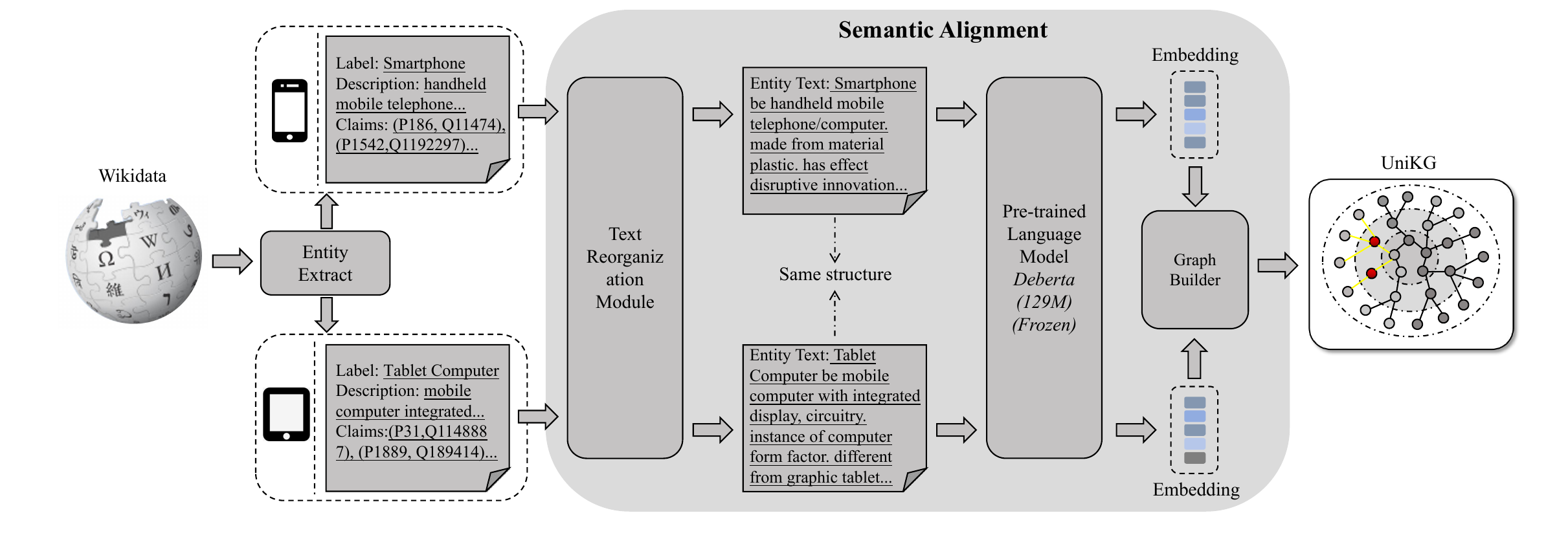}
	\caption{Overall architecture of heterogeneous graph construction. We use smartphones and tablet computers as examples to describe the architecture. Firstly, we extract and filter the attributes of these two entities from Wikidata. These attributes are summarized into two paragraphs with similar structures through the Text Reorganization Module. Their embeddings are reasoned through a PLM. Finally, the directed edges with multiple attributes are built through Graph Builder. The positions and links of these two entities are highlighted in UniKG.}
	\label{fig:example_of_extration}
\end{figure*}

Although notable success has been achieved by existing works on HGs, there are still several issues to be tackled. The most important one is the lack of a large-scale HG dataset. Existing HG datasets are almost small-scale, with just thousands of nodes and dozens of relation types limited in specific knowledge domains. This greatly limits the ability to facilitate representation learning and knowledge extraction, therefore there is an urgent need for a large-scale HG dataset with encyclopedic knowledge. Furthermore, when regarding effective learning on large-scale HG datasets, another issue is the inapplicability of existing HG learning methods. On one hand, existing aggregation-based HG learning methods cost too much memory overhead when performing multi-hop aggregation in very large-scale receptive fields. Hence, they may suffer from out-of-memory problems during model training. On the other hand, for the meta-path-based methods, existing works are applied to the case where only one type of association exists between two types of nodes. However, when the association types between two nodes significantly increase, the number of metapaths may suffer explosive growth (Please see the example in Figure~\ref{fig:pathmethod}). This causes great difficulty for topology perception if only a limited number of metapaths can be sampled.

In this paper, we construct a large-scale Heterogeneous Graph benchmark dataset named UniKG from Wikidata.
We release UniKG to facilitate the representation learning task of heterogeneous graphs.
Overall, UniKG contains 77.31 million multi-attribute entities labels by 2000 classes, 564 million directed edges annotated by 2082 diverse association types, which significantly surpasses the scale of existing homogeneous graph datasets.
To perform effective learning on the large-scale UniKG, two key measures are taken.
We propose a semantic alignment strategy for multi-attribute entities, which projects the feature description of multi-attribute nodes into a common embedding space.
Then we design a novel plug-and-play anisotropy propagation module (APM) as base to extend methods of large-scale homogeneous graphs to heterogeneous graphs.
These two strategies enable efficient information propagation among a tremendous number of multi-attribute entities and meantimes adaptively mine multi-attribute association through the multi-hop aggregation in large-scale HGs.
We set up a node classification task on our UniKG dataset that evaluate multiple baseline methods which are constructed by embedding our APM into large-scale homogenous graph learning methods.
In addition, we attempt to transform knowledge of UniKG to a recommendation system to verify the capability to facilitate downstream task of diverse domains.

Our contributions can be summarized in the following three aspects:

1) We construct a large-scale Heterogeneous Graph benchmark dataset named UniKG from Wikidata which contains over 77 million nodes and 564 million edges associated with 2,082 types.
To the best of our knowledge, UniKG significantly surpasses the scale of existing HG datasets.

2) We propose a plug-and-play Anisotropy Propagation Module (APM), which can adaptively mine association relationships of diverse types and extends methods of large-scale homogeneous graphs to heterogeneous graphs.

3) We transform knowledge from UniKG to recommender systems, which effectively promotes performance of recommendation task.

\begin{table*}[t]
	\centering
	\caption{Statistics of proposed UniKG dataset and common heterogeneous node classification datasets.
		It can be observed that the scale of UniKG significantly surpasses the scale of others.}
	\begin{tabular}{cccccc} \hline
		\toprule
		Datasets   & \#Nodes    & \#Node Types & \#Edges    & \#Edge Types & \#classes \\
		\midrule
		ACM        & 10,942      & 4          & 273,936    & 8          & 3 \\
		IMDB       & 21,420      & 4          & 43,321      & 6          & 5 \\
		DBLP       & 26,128      & 4          & 119,783    & 6          & 4 \\
		Freebase   & 180,098     & 8          & 528,844    & 36         & 7 \\
		MAG        & 1,939,743    & 4          & 21,111,007   & 4       & 349 \\
		OAG        & 1,116,162    & 5          & 13,965,692 & 30         &  275 \\
		UniKG      & 77,312,474   & 2,000     & 564,425,621  & 2,082       & 2,000 \\
		\bottomrule
	\end{tabular}%
	\label{tab:Dataset_Statistics}%
\end{table*}%

\section{Related Work}

In this section, we introduce the classical HG datasets, as well as representative GNN methods, such as those for solving KGs, large-scale graphs and HGs tasks.
Finally, we briefly discuss the limitations of existing datasets and methods and point out highlights the advantages of our proposed datasets and methods.

\subsection{Heterogeneous Graph datasets}

Numerous studies have focused on constructing domain-specific heterogeneous graph datasets \cite{sharma2022finred, chen2023knowledge, ahrabian2023pubgraph, jang2022temporalwiki} to facilitate research in their respective fields.
For instance, \cite{sharma2022finred} demonstrate the insufficiency of existing relation extraction tasks in the financial domain.
The recently proposed PubGraph~\cite{ahrabian2023pubgraph} contains knowledge from almost all scientific fields.
\cite{jang2022temporalwiki} has created a temporal benchmark to address temporal misalignment issues.
Moreover, some studies~\cite{stranisci2022urw,kubevsa2023damuel} attempt to explore cultural differences by extracting and combining diverse linguistic entities or events.
For example, \cite{stranisci2022urw} reveals the issue of underrepresentation of non-Western writers.

\begin{figure*}[ht] 
	\centering
	\includegraphics[width=1\linewidth]{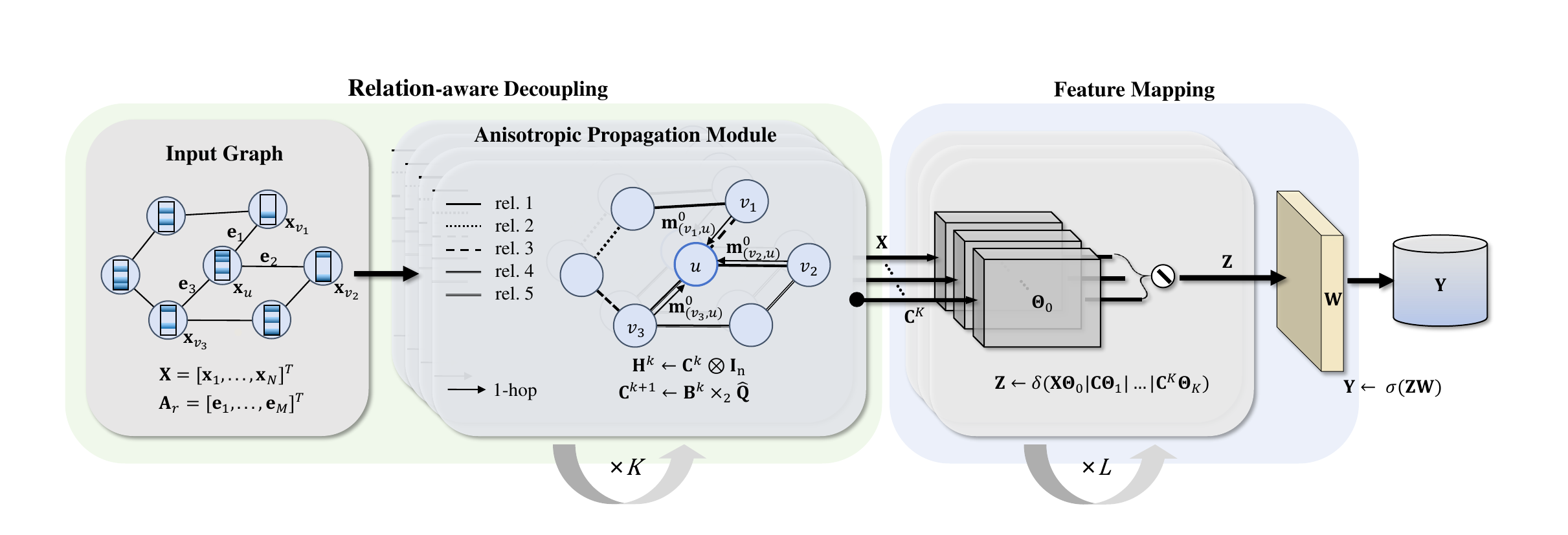} \vspace{-0.5cm}
	\caption{Overall architecture of the Anisotropic Graph Decoupling.
		We use node $u$ as an example to describe our APM, in the diagram, node $u$ is associated with three nodes $v_1$, $v_2$, $v_3$.
		In single-hop propagation of APM, the node $u$ captures the semantic structure in an edge-aware manner.
		Structures with multiple attribute associations generate anisotropic propagation features of $K$+1 hops from $\mathbf{X}$ to $\mathbf{C}^{K}$ through the $K$ hop propagations of APM.
		Then, the $K$+1 hop propagation features are input into the post classifiar to generate the predicted label matrix $\mathbf{Y}$.
	}
	\label{fig1:architecture}
\end{figure*}

\subsection{Graph Neural Networks Methods}

As existing work is unsuitable for large-scale heterogeneous graphs, we review GNN methods designed for KG tasks, large-scale graphs and HGs.

\textbf{GNNs Methods for KGs tasks}
The similar structure has led to the increased use of GNN methods for KG tasks.
For instance, \cite{banerjee2023gett} proposes a question-answering system GETT-QA based on GNNs embeddings.
\cite{pramanik2021uniqorn} constructs local context graphs for information retrieval.
Recently, \cite{luo2023dhge} uses two branches of Hyper-graph Neural Networks DHGE to learn entity instance views and concept ontology views.
One particularly interesting research is KGQ~\cite{hogan2021knowledge}, which discusses a unified data model derived from KGs and queries, which coincides with our idea.
\textbf{GNNs Methods for Large-Scale Graphs}
Methods specifically designed for large-scale graphs are commonly categorized into two groups: sampling-based aggregation GNNs and decoupling-based GNNs.
Sampling-based aggregation GNNs sample subgraphs by various strategies, such as node sampling~\cite{hamilton2017inductive}, neighborhood sampling~\cite{chen2018fastgcn}, and subgraph sampling~\cite{chiang2019cluster,zeng2019graphsaint}, seeking the optimal way to perform batch training.
Instead, decoupling-based GNNs precompute the feature propagation matrix once and store the results as inputs for the models.
For instance, \cite{wu2019simplifying} proposes SGC, which uses the last-hop propagation feature as input.
Then, SIGN~\cite{frasca2020sign} concatenates multi-hop propagation features, and SAGN~\cite{sun2021scalable} further introduces a hop-wise attention mechanism.
\textbf{GNNs Methods for Heterogeneous Graphs}
HGNNs face a major obstacle in dealing with the heterogeneity arising from complex structures.
An instance is PME~\cite{chen2018pme}, which measures the distances between nodes to capture heterogeneity.
\cite{ma2021heterogeneous} proposes GATNE, aiming to learn node embedding in multi-path graphs.
Recently, based on the attention mechanism, scholars proposed a series of attention-based HGNNs~\cite{fu2020magnn,hong2020attention,hu2020heterogeneous,fu2019metapath} to capture semantic importance of nodes.

In contrast, our dataset is large-scale and universal domain, which can support model training on a large-scale heterogeneous graph.
And our method can adaptively mine multi-attribute association relationships with low complexity, and extend methods of large-scale homogeneous graphs to heterogeneous graphs.

\section{Preliminary}
\subsection{Heterogeneous Graph}
A Graph can be defined as $ \mathcal{G} = {(\mathcal{V}, \mathcal{E})} $, in which $ \mathcal{V} $ and $ \mathcal{E} $ represent the sets of nodes and edges, respectively.
Each node $ v_i \in \mathcal{V} $ and each edge $ e_i \in \mathcal{E} $ are associated with a only one specific type.
Formally, a heterogeneous graph (or heterogeneous information network) can be defined as follows:
\[
\mathcal{G} = (\mathcal{V}, \mathcal{E}, \mathcal{A},\mathcal{R}, \phi_v, \phi_e)
\]
where:
\begin{itemize}
\item $\mathcal{V} = \{v_1, v_2, \ldots, v_N\}$ denotes the set of nodes, and $N$ is the total number of nodes.
\item $\mathcal{E} = \{e_{ij}\}$ denotes the set of directed edges between nodes.
\item $\mathcal{A} = \{a_1, a_2, \ldots, a_Q\}$ is the set of node classes, and $Q$ is the total number of node types.
\item $\mathcal{R} = \{r_1, r_2, \ldots, r_M\}$ is the set of edge types, and $M$ is the total number of edge types.
\item $\phi_v: \mathcal{V} \rightarrow \{a_1, a_2, \ldots, a_Q\}$ represents the node labeling function that maps nodes to their corresponding classes.
\item $\phi_e: \mathcal{E} \rightarrow \{r_1, r_2, \ldots, r_M\}$ represents the edge type function that maps edges to their corresponding types.
\end{itemize}

\subsection{Decoupled Methods on Large-Scale Homogeneous Graphs}

Given a graph $ \mathcal{G}=(\mathcal{V},\mathcal{E}) $, Each node $ v_{i} \in \mathcal{V} $ is embedded as a $ d_v $-dimensional vector $ \mathbf{x}_i $, forming the node feature matrix $ \mathbf{X} $. Each directed edge $ e_i \in \mathcal{E} $ is a [0,1] scalar value, forming adjacent matrix $\mathbf{A} \in \mathbb{R}^{n\times n}$.
The decoupled methods on large-scale homogeneous graphs are proven effective paradigm for large-scale graph representation learning.
They iteratively aggregate propagation features to update embeddings, which can be formulated as:
\begin{equation}
	\begin{cases}
		\mathbf{Z}=\delta([\mathbf{X}\mathbf{\Theta}_{0}|\mathbf{P}\mathbf{X}\mathbf{\Theta}_{1}|...|\mathbf{P}^{K}\mathbf{X}\mathbf{\Theta}_{K}]) \\
		\mathbf{Y}=\sigma(\mathbf{Z} \mathbf{W})
	\end{cases}
\end{equation}

where $K$ refers to the maximum number of hops, $\mathbf{Z}$ denotes the hidden variable, $ \mathbf{P}^{k}\mathbf{X} $ refers to $ k $-th hop propagation matrix and $ \mathbf{P}=\mathbf{D}^{-1/2}\mathbf{A}\mathbf{D}^{-1/2}$ is the symmetric regularization matrix, $ \delta(\cdot) $ represents the aggregation function to fuse multi-hop features (e.g. sum$(\cdot)$, mean$(\cdot)$, or an attention mechanism), $\sigma(\cdot)$ is the non-linear activation function, $ \mathbf{\Theta} $ and $ \mathbf{W} $ are learnable parameter matrices.
Based on the widely accepted consensus that messaging passing overheads much more, the decoupling-based GNNs precompute the approximate result of the message passing process in the CPU as the input of overall model train.
Thus, this architecture is simple and efficient.

\section{Dataset Construction}

In this section, we introduce our proposed dataset UniKG, including construction details and data statistics.
Figure~\ref{fig:example_of_extration} is a construction example.
Table~\ref{tab:Dataset_Statistics} and Table~\ref{tab:Data_Split} show the statistics and three different scales of UniKG, respectively.

\subsection{Entities Extraction}
Most knowledge graphs, such as Wikidata, have an underlying database model that provides various functions for organizing data of a specific type.
Directly deserializing the data of billions of entities to construct a large-scale graph is very difficult.
Thus, we design an extraction strategy to carefully parse Wikidata and automatically filter out entities with complete attributes (e.g., contain `id', `description', `labels', and `claims' in English).
For example, when parsing entity `Belgium', we extract its id as `Q31', description is `country in Western Europe', label is `Belgium', and dozens of `claims' triplets.
In addition, to ensure the semantic adequacy of the filtered entities, we semi-manually generated a screen set to filter the mass of unsemantic `claims' attributes in Wikidata.
For example, we found that Wikidata stores lots of `external-id' relationships, which lack semantics and are thus ignored.
The <subject, object> structure of filtered claims can be modeled as directed edges between entities.
In this way, we extract over 77 million nodes and 564 million directed edges.

\subsection{Semantic Alignment}

To facilitate node aggregation in a large receptive field, we design a alignment strategy to contain significant semantic structure.
Specifically, the first sentence of each entity is organized into the structure of $<$label `be' description.$>$, which describes the attributes of the entity itself.
Then, for each filtered `claim', its label $c$ and the associated entity label or value $e$ are organized into the structure $<c$ $e$.$>$, which is sequentially linked to first sentence.
In this way, each entity is automatically organized into a feature description that conforms to a significant structure.
We randomly sampled a portion of entities to fine-tune the Pre-trained Language Model (PLM) \textit{Deberta}.
Then, the PLM projects feature description of all entities into a common representation space to generate node embedding.

\subsection{Label Annotation}

Entities in Wikidata are redundantly and trivially labeled, such as `fictional character', `character that may or may not be fictional', `character in a fictitious work', `conceptual character', `fictional character in a musical work', and `imaginary character', which results in similar entities may have multiple different annotation.
This is not favorable for classification tasks and is a heavy burden for representation learning tasks.
Therefore, we design an annotation strategy to semi-automatically annotate all filtered entities. 
i) For each entity $e \in \mathcal{E}$, we extract all `instance of' relationships.
These relationships link to the parent entity set $\mathcal{F}(a)=\{b_1,b_2...,b_f\}$, where $b$ associate each `instance of' and $f$ represents the number of `instance of'.
The parent entity sets of all entities are collected into a union as the label set $\mathcal{F}_{all}=\{\bigcup\mathcal{F}(a)|a\in\mathcal{E} \}$.
ii) Then, we cluster (e.g. k-means) all of the feature description embeddings of $\mathcal{F}_{all}$, resulting in a smaller high-level label set $\mathcal{F}_m$.
For instance, we annotate the labels `fictional character', `character that may or may not be fictional', `character in a fictitious work', `conceptual character', `fictional character in a musical work', and `imaginary character' as `Fictional characters from various mediums'.
Finally, we annotate each entity with the help of TWO non-native fluent English speakers, who fine-checked the incorrect associations between specific labels and high-level labels.
All associations marked as incorrect by both the annotators are manually relabeled.
Additionally, each directed edge is annotated as associated relationship label.
In this way, we cluster 74666 `instances of' object entities into 2000 abstract manual annotations with physical meaning, and obtain 564 million edges with 2,082 relationship labels.

Through these strategies, we obtain 77,312,474 nodes labeling by 74,666 semi-manual annotations and 564,425,621 directed edges of 2,082 distinct types, and construct an incredibly intricate large-scale heterogeneous graph.
In previous heterogeneous graphs, an edge usually has only one type, however, in our dataset an edge may have multiple types.
For example, `member of' and `part of', `author of' and `creator of ' tend to co-exist in a node-pair.
This results in metapaths may suffer explosive growth.

\begin{table}[t]
	\centering
	\caption{Statistics of different scales of UniKG.}
	\begin{tabular}{cccc} \hline
		\toprule
		Datasets   & \#Nodes & \#Edges  & \#Classes \\
		\midrule
		UniKG-1M   & 1,002,988     & 24,475,405 & 2,000 \\
		UniKG-10M  & 10,044,777    & 216,295,022 & 2,000 \\
		UniKG-full & 77,312,474    & 564,425,621 & 2,000 \\
		\bottomrule 
	\end{tabular}%
	\label{tab:Data_Split}%
\end{table}%

\begin{table*}[ht]
	\centering
	\caption{Performance of the five baselines on three scale UniKGs. `R-' means use our HGD framework.}
	\begin{tabular}{c|cccc|cccc|cccc}
		\toprule
		\multirow{2}[4]{*}{Methods} & \multicolumn{4}{c|}{UniKG-1M}                     & \multicolumn{4}{c|}{UniKG-10M}                    & \multicolumn{4}{c}{UniKG-Full} \\
		\cmidrule{2-13}               & Acc.       & Prec.      & Rec.       & F1.        & Acc.       & Prec.      & Rec.       & F1.        & Acc.       & Prec.      & Rec.       & F1. \\
		\midrule
		R-MLP & 68.48 & 92.24 & 68.31 & 78.50 & 78.66 & 96.77 & 79.60 & 87.35 & 80.69 & 97.25 & 81.69 & 88.79 \\
		R-SGC & 44.18 & 88.47 & 42.75 & 57.65 & 64.25 & 93.00 & 65.21 & 76.67 & 69.84 & 94.79 & 71.40 & 81.45 \\
		R-SIGN & 66.69 & \textbf{94.35} & 65.27 & 77.16 & 69.52 & \textbf{96.84} & 70.18 & 81.38 & 75.45 & 97.58 & 76.42 & 85.71 \\
		R-SAGN & \textbf{75.41} & 90.64 & \textbf{75.95} & \textbf{82.64} & \textbf{89.03} & 96.19 & \textbf{90.11} & \textbf{93.05} & \textbf{93.16} & \textbf{98.46} & \textbf{93.83} & \textbf{96.09} \\
		R-GAMLP & 47.24  & 86.95 & 46.80 & 60.85 & 60.99 & 88.44 & 62.70 & 73.38 & 72.89 & 94.28 & 74.95 & 83.51 \\
		\bottomrule
	\end{tabular}%
	\label{tab:exp1}%
\end{table*}%
\begin{figure*}[ht] 
	\centering
	\includegraphics[width=1\linewidth]{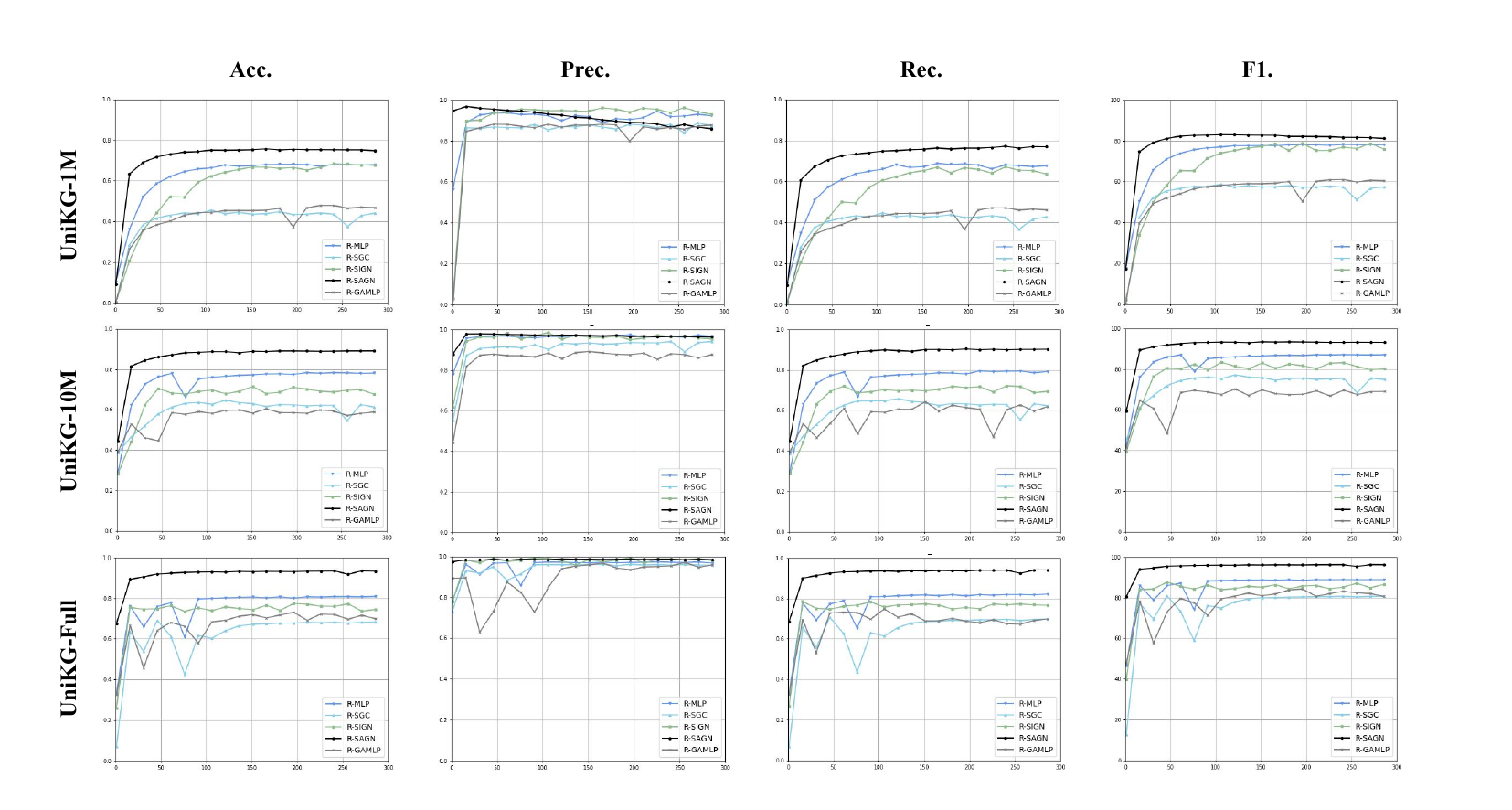} 
	\caption{Subset Accuracy, Precision, Recall and F1 score of five baselines on three scales of UniKG.
	}
	\label{fig1:allfigs}
\end{figure*}

\section{Anisotropic Graph Decoupling}

In this section, we detailed describe the novel Heterogeneous Graph Decoupling (HGD) framework, and the anisotropic propagation process of the plug-and-play Anisotropic Propagation Module (APM) on large-scale heterogeneous graphs as shown in Figure~\ref{fig1:architecture}.

Through introducing $ d_e $-dimensional edge embeddings, the adjacency matrix $ \mathbf{A}\in \mathbb{R}^{n\times n} $ can be represented as a relation-aware adjacency matrix $ \mathbf{A}_{r}\in \mathbb{R}^{n\times n\times d_e} $.
Therefore, the Heterogeneous Graph Decoupling (HGD) framework can be represented as:
\begin{equation} 
	\begin{cases}
		\mathbf{Z}=\delta([\mathbf{X}\mathbf{\Theta}_{0}|\mathbf{C}\mathbf{\Theta}_{1}|...|\mathbf{C}^{K}\mathbf{\Theta}_{K}]) \\
		\mathbf{Y}=\sigma(\mathbf{Z} \mathbf{W})
	\end{cases}
\end{equation} 
where, for each $k \in [1, K]$, $\mathbf{C}^{k} \in \mathbb{R}^{n\times d}$ denotes the anisotropic propagation matrix of the $k$-th hop.
We formulate the process of anisotropic propagation as a function $\hat{f}(\cdot)$, then \(\mathbf{C}^{k+1} = \hat{f}(\mathbf{A}_r, \mathbf{C}^k) \) and \( \mathbf{C}^0 = \mathbf{X} \) is the original feature matrix.

To integrate the anisotropic edge features into the message matrix, we first upgrade dimension $\mathbf{C}^{k}$ from $\mathbb{R}^{n \times n}$ to $\mathbb{R}^{n \times n \times d}$, and the matrix is denoted as:
\begin{equation}
	\mathbf{H}^{k}=\mathbf{C}^{k} \otimes \mathbf{I}_n
	\vspace{-0.03cm}
\end{equation}
where \( \mathbf{I}_n \in \mathbb{R}^{n \times n} \) is the identity matrix.
The \(i\)-th row of \( \mathbf{H}^{k} \) represents a message matrix \( \mathbf{M} \in \mathbb{R}^{n \times d} \) from node $i$ to others. Based on the $\mathbf{H}^{k}$, through the tensor product of the anisotropic message matrix and its coefficient matrix, we can obtain the anisotropic propagation matrix at the $k+1$-th hop by the attention mechanism, which can be expressed as:
\begin{equation}
	\mathbf{C}^{k+1} = \mathbf{B}^{k} \times_2 \hat{\mathbf{Q}}
\end{equation}
where $\times_2$ represents the tensor product on the second dimension, \( \mathbf{C}^{k+1} \in \mathbb{R}^{n \times d} \) denotes the \( (k+1) \)-th hop anisotropic propagation matrix, $\mathbf{B}^{k}$ is the message propagation matrix, and $\hat{\mathbf{Q}}$ represents the edge-aware coefficient matrix.
Specifically, we perform element-wise Hadamard multiplication $\odot$ between \( \mathbf{A}_r \) and \( \mathbf{H}^{k} \) to derive the message propagation matrix \( \mathbf{B}^{k} \in \mathbb{R}^{n \times n \times d} \):
\begin{equation}
	\mathbf{B}^{k} = \mathbf{A}_r \odot \mathbf{H}^{k} 
\end{equation}
where, $\mathbf{B}^{k}$ describes messages between each node-pairs.
The \( l_2 \)-norm matrix of \( \mathbf{B}^{k} \) on the third dimension can be represented as $\mathbf{Q} \in \mathbb{R}^{n \times n}$:
\begin{equation}
	\vspace{-0.03cm}
	\mathbf{Q} = ||\mathbf{B}^{k}||_2
	\vspace{-0.03cm}
\end{equation}
then, calculate the edge-aware coefficient matrix \( \hat{\mathbf{Q}} \in \mathbb{R}^{n \times n} \) for the messages between each node-pair by row-wise softmax:
\begin{equation}
	\vspace{-0.03cm}
	\hat{\mathbf{Q}} = \text{softmax}(\mathbf{Q})
	\vspace{-0.03cm}
\end{equation}
Finally, perform tensor multiplication along the second dimension between the message propagation matrix \( \mathbf{B}^{k} \) and the edge-aware coefficient matrix \( \hat{\mathbf{Q}} \), which means that messages of $k$-th hop are anisotropically propagate to the corresponding nodes of $k+1$-th hop.
By iterating this procedure \( K \) times, we obtain a sequence of anisotropic propagation matrices $[\mathbf{C}^{1}, ..., \mathbf{C}^{K} ]$, which serves as the input to the post-classifier of the decoupling-based graph neural network.

This method supports fast multi-attribute aggregation and can extends decoupled methods of the large-scale homogeneous graphs to the large-scale heterogeneous graphs in a edge-aware manner.

In real-world, nodes tend to be annotated by multi-label.
We use multi-label loss functions, such as the Binary Cross-Entropy (BCE) loss:
\begin{equation}
	\mathcal{L} = -\frac{1}{N} \sum_{i=1}^{N} \left[ \mathbf{y}_i \cdot \log(\hat{\mathbf{y}}_i) + (1 - \mathbf{y}_i) \cdot \log(1 - \hat{\mathbf{y}}_i) \right]
\end{equation}
where $N$ is the number of nodes, $\mathbf{y}_i$ denotes the ground truth vector of node $i$, and $\hat{\mathbf{y}}_i$ represents the predicted label vector.

\begin{table*}[ht]
	\vspace{-0.1cm}
	\centering
	\caption{Results of ablation experiment on recommendation system. Where the `w/ u.' means `with prior knowledge from UniKG-Full'.}
	\begin{tabular}{c|ccc|ccc|ccc} \hline
		\toprule
		\multirow{2}[4]{*}{Methods} & \multicolumn{3}{c|}{Amazon-book}     & \multicolumn{3}{c|}{Yelp}       & \multicolumn{3}{c}{Citeulike-a} \\
		\cmidrule{2-10}               & Prec.      & Rec.       & Ndcg.      & Prec.      & Rec.       & Ndcg.      & Prec.      & Rec.       & Ndcg. \\
		\midrule
		LightGCN   & 0.01716    & 0.06191     &0.04106 & 0.00433    & 0.01123    & 0.00849     & 0.02329    & 0.07188    & 0.05064 \\
		LightGCN w/ u. & 0.52\%    & 0.11\%    & 0.82\%     & 6.47\%    & 7.93\%    & 4.36\%    & 0.09\%    & 0.26\%    & 0.55 \% \\
		\bottomrule
	\end{tabular}%
	\label{tab:exp2}
\end{table*}%

\begin{table}[ht]
	\centering
	\caption{Dataset statistics for recommendation systems. Where `Cul.' means `Citeulike-a', `Ama.' means `Amazon-book'.}
	\begin{tabular}{m{3em}<{\centering}m{3em}<{\centering}m{3.3em}<{\centering}m{4em}<{\centering}m{4.3em}<{\centering}} \hline
		\toprule
		Datasets   & \#Users      & \#Items      & \#Edges & \#Relations \\
		\midrule
		Cul. & 5,551       & 16,980      & 210,537     & 1 \\
		Ama. & 52,643      & 65,865      & 2,090,149    & 39 \\
		Yelp & 44,907      & 137,597     & 2,346,409    & 42 \\
		\bottomrule
	\end{tabular}%
	\label{tab:re_stat}%
	\vspace{-0.5cm}
\end{table}%

\section{Experiments}

In this section, we introduce two experiments to evaluate the proposed dataset and method, along with the ability to facilitate other downstream tasks.
The detailed descriptions of recommendation task are reported in Appendices.

\subsection{Experimental Setup}

\paragraph{Datasets}
In Table~\ref{tab:Dataset_Statistics}, we compared UniKG with common heterogeneous node classification datasets.
It can be observed that UniKG significantly surpasses the scale of existing HG datasets in terms of the numbers of nodes, edges, edge types, etc.
Due to the high complexity associated with the enormous size of UniKG-Full, we construct two sub-datasets of small and medium sizes to accommodate different scales of models, through Snowball Sampling~\cite{goodman1960snowball} following~\cite{ahrabian2023pubgraph}.
Specifically, we extract high-degree nodes and random nodes as the start of random walk to traverse and record 1 million and 10 million nodes. 
Considering sample bias, we conduct full-graph anisotropic propagation before the split.
Table~\ref{tab:Data_Split} shows the statistics of three different scales.
We set up and evaluate a semi-supervised multi-label node classification task on the UniKG-1M, UniKG-10M, and UniKG-Full datasets.
Each dataset is randomly split into training, validation, and testing sets using an 8:1:1 ratio.
Moreover, we conduct the recommendation task ablation experiment on the Amazon-Book, Yelp and Citeulike-a datasets.
Their statistics are shown in Table~\ref{tab:re_stat}.
\vspace{-0.1cm}

\paragraph{Baselines}
For the node classification task, incorporating \textbf{MLP}, \textbf{SGC}~\cite{wu2019simplifying}, \textbf{SIGN}~\cite{frasca2020sign}, \textbf{SAGN}~\cite{sun2021scalable}, and \textbf{GAMLP}~\cite{zhang2022graph} into our HGD frame work, we propose 5 baselines: \textbf{R-MLP}, \textbf{R-SGC}, \textbf{R-SIGN}, \textbf{R-SAGN}, and \textbf{R-GAMLP} that are applicable to large-scale heterogeneous graphs.
For another recommendation task, we adopt \textbf{LightGCN}~\cite{he2020lightgcn} as our baseline model.
\vspace{-0.1cm}
\paragraph{Ablations}
For the recommendation task, we use similar embeddings retrieved in UniKG as more expressive initial representations rather than random initialization.
Thus, the initial representations from UniKG-Full datasets can form 2 ablation environment: LightGCN, and LightGCN w/ UniKG-Full for ablation study.
\vspace{-0.1cm}
\paragraph{Evaluation Metrics}
For the multi-label node classification task, we choose Accuracy, Precise, Recall, F1 Score as evaluation indicators.
Note that we choose `Subset Accuracy' as Accuracy, which is the proportion of nodes whose predicted labels exactly match the ground truth labels.
For the recommendation task, we choose Precise, Recall and Ndcg as evaluation indicators.
\vspace{-0.1cm}
\paragraph{Hyperparameters and Environment}
Certain reported performances on the OGB leaderboard~\cite{hu2020open} have incorporated various tricks such as SLE, RLU, and the `labels as input'~\cite{wang2021bag}.
These tricks are recognized as a significant influence on performance~\cite{sun2021scalable}.
To ensure fairness, we avoid these tricks completely.
Furthermore, we refer to the hyperparameter settings of the official OGB~\cite{hu2020open} or the original authors.
For the node classification task, we adopt the following hyperparameters for all models: embedding dimension = 128 for both entities and relations, hidden dimension = 256, epoch = 300, learning rate = 0.01, and dropout rate = 0.
For the recommendation task, we adhere to the original configuration of the LightGCN.
All experiments were conducted using single 24GB GeForce GTX 4090 GPU.
\vspace{-0.1cm}

\subsection{Main Results and Analysis}
In this experiment, we evaluate all baselines on UniKG and compare ablations on recommendation task.

\paragraph{Node Classification Task}
From Table~\ref{tab:exp1}, we observe a clear trend: as the ability to capture multi-hop features improves, performance gradually enhances.
This trend aligns with conclusions drawn from existing large-scale graph benchmarks.
Specifically, R-SAGN consistently outperforms other baselines, achieving 93.16\% subset accuracy and 96.09\% F1 score on the largest UniKG-Full dataset.
We attribute advantage of SAGN to its adaptive attention mechanism, which effectively aggregates multi-attributes multi-hop propagation features.
This may imply that it is crucial to design post classifier with heterogeneous graph structure-aware components.
Additionally, we find that R-MLP, utilizing only the original features, achieves a subset accuracy of 80.69\%.
In contrast, R-SGC, utilizing only the last-hop propagation features, decreased by 10.85\%, R-SIGN, using multi-hop propagation features, decreased by 5.24\%, and GAMLP, using the layer-wise attention mechanism, decreased by 7.80\%.
Similar trends were observed on the other two smaller datasets.
In summary, for the representation learning task on large-scale heterogeneous graphs, we conclude that:
i) The Heterogeneous Graph Decoupling framework proved to be effective;
ii) The original features are important, so it is necessary to increase their importance or incorporate residual structures;
iii) In the feature mapping process, it is crucial to design effective aggregation strategies, such as the adaptive attention mechanism in R-SAGN, to capture heterogeneous multi-hop propagation features.

\paragraph{Recommendation Task}
The similar embeddings retrieved in UniKG as the more expressive initial embedding in the recommender system and are used as the ablation condition.
Table~\ref{tab:exp2} presents the results of ablation experiments.
It can be observed that the performance of LightGCN improves on all datasets after using the more expressive initial embedding.
In particular, on the Yelp dataset, Precision improves by 6.47\%, Recall by 7.93\%, and Ndcg by 4.36\%.
This indicates that more expressive features can improve model performance and that our proposed UniKG can facilitate downstream tasks of recommender systems.

\vspace{-0.2cm}
\section{Conclusion}

We construct a large-scale heterogeneous graph benchmark dataset UniKG, along with associated representation learning methods: the Heterogeneous Graph Decoupling framework and a plug-and-play Anisotropy Propagation Module.
Our proposed construction methods model knowledge graphs as large-scale heterogeneous graphs, and the scale of the constructed UniKG significantly surpasses the scale of existing heterogeneous graph datasets.
The integration of the decoupling-based methods of homogeneous graphs with Heterogeneous Graph Decoupling framework is easy to implement and applicable to large-scale heterogeneous graphs with competitive performance. 
The real-world knowledge of UniKG can facilitate diverse downstream tasks.

\bibliography{aaai24}

\end{document}